\documentclass[a4paper]{article}

\usepackage{INTERSPEECH2022}

\usepackage{amsmath,graphicx}
\usepackage{booktabs}
\usepackage{makecell}
\usepackage{subcaption}
\usepackage{enumitem}
\usepackage{etoolbox}
\usepackage{xcolor}
\usepackage{amsfonts}

\def\vomega{{\boldsymbol \omega}}
\def\vg{{\boldsymbol g}}
\def\vv{{\boldsymbol v}}
\def\vG{{\boldsymbol G}}
\def\vtheta{{\boldsymbol \theta}}

\title{Online Continual Learning of End-to-End Speech Recognition Models}
\name{Muqiao Yang, Ian Lane, Shinji Watanabe
\thanks{
This work used the Extreme Science and Engineering Discovery Environment (XSEDE) ~\cite{xsede}, which is supported by National Science Foundation grant number ACI-1548562. Specifically, it used the Bridges system ~\cite{nystrom2015bridges}, which is supported by NSF award number ACI-1445606, at the Pittsburgh Supercomputing Center (PSC).}}
\address{Carnegie Mellon University, Pittsburgh, PA, USA}
\email{\{muqiaoy, ianlane, swatanab\}@andrew.cmu.edu}

\begin{document}

\maketitle
\begin{abstract}
  Continual Learning, also known as Lifelong Learning, aims to continually learn from new data as it becomes available. While prior research on continual learning in automatic speech recognition has focused on the adaptation of models across multiple different speech recognition tasks, in this paper we propose an experimental setting for \textit{online continual learning} for automatic speech recognition of a single task. Specifically focusing on the case where additional training data for the same task becomes available incrementally over time, we demonstrate the effectiveness of performing incremental model updates to end-to-end speech recognition models with an online Gradient Episodic Memory (GEM) method. Moreover, we show that with online continual learning and a selective sampling strategy, we can maintain an accuracy that is similar to retraining a model from scratch while requiring significantly lower computation costs. We have also verified our method with self-supervised learning (SSL) features.
\end{abstract}

\section{Introduction}
\label{sec:intro}



Recent advances in machine learning have enabled systems to approach or even surpass the performance of humans on some specific tasks \cite{xiong2016achieving}. However, existing machine learning models are typically trained on a fixed set of data. When additional training data becomes available the model needs to be retrained from scratch in order to achieve the best accuracy possible with the given data. For each update, this process takes significant time and computational resources, which is environmentally unfriendly due to carbon dioxide emissions. Alternatively, if the model is updated incrementally using only the new data (which requires significantly less computational resources) the model may Catastrophic Forget \cite{mccloskey1989catastrophic}, degrading the overall accuracy.  

Research on continual learning, or lifelong learning, is focused on methods to overcome the problem of catastrophic forgetting. These methods accumulate new and past knowledge in an efficient way enabling models to operate at higher and higher accuracy as additional training data becomes available \cite{chen2018lifelong}. Many continual learning methods get inspiration from biology, as humans have the ability to continually learn new knowledge and skills based on their prior experience and apply it to later scenarios \cite{bremner2012multisensory}. From a biological perspective, the lifelong learning mechanisms in the brain are modular \cite{bremner2012multisensory}. Machine learning methods like \cite{rusu2016progressive, guo2020continual} use isolated parameters when learning a new task and expand the network architecture during the learning process, mimicking this modular structure. Other methods \cite{lopez2017gradient, chaudhry2019continual, d2019episodic} store data from the past and replay it while learning a new task. These approaches are inspired by the fact that memory mechanisms help humans retain the knowledge of past experiences so that they can learn to consolidate both past and new knowledge over a lifetime \cite{tulving19721o}. 

Most existing research in continual learning of machine learning models has focused on tasks such as computer vision and reinforcement learning \cite{chen2018lifelong, aljundi2017expert, shin2017continual}. There have been limited studies on the effectiveness of continual learning of automatic speech recognition (ASR) and prior research in this area has focused on updating ASR models to perform well on new tasks or domains \cite{sadhu2020continual, chang2021towards, xue2019multi, fu2020incremental, houston2020continual, kessler2021continual}. In many real-world applications, however, we encounter the scenario of an ASR model being used for a product (such as voice transcription, voice search, or conversational interaction) where the application task remains the same over time, but there are continuous streams of data that are available to periodically collect and label. This data can potentially be used to continually improve the model accuracy over time. In most cases, the amount of new data that becomes available will be significantly smaller than the data used to train the original model, and the underlying data distributions might be non-stationary \cite{zeno2018task}. This could make it difficult to update the model for traditional continual learning approaches as most of them would require access to explicit task boundaries between domains. Moreover, when the model is getting access to new data frequently, it would be time-consuming and impractical to go over each of the previous tasks and update the model every time for the new data. 

Different from existing continual learning methods in ASR, in this paper, we will focus on the online scenario for continual learning of end-to-end speech recognition models, where continuous streams of speech become available over time and there are no explicitly defined task boundaries, and the small sets of new unseen data are incoming in an online manner. Besides, one of our technical novelties is to show that performing incremental model updates on the TED-LIUM corpus \cite{rousseau2012ted, rousseau2014enhancing, hernandez2018ted} with an online Gradient Episodic Memory (GEM) method can achieve a better trade-off between model performance and computational cost in comparison to other techniques. Also, we propose to apply GEM with various sampling strategies to this unique online learning setup in speech recognition and show that a selective sampling mechanism would help the model converge better. At last, we also demonstrate that our continual learning methodology has compatibility with the latest self-supervised learning models and other similar datasets.




\section{Related Work}
\label{sec:related}


Continual learning methods can be divided into three main categories: regularization approaches, architectural approaches, and replay-based approaches. Regularization approaches introduce an additional regularization term when new tasks are encountered in order to adjust parameter importance \cite{kirkpatrick2017overcoming, zenke2017continual, aljundi2018memory, li2017learning}. Architectural approaches isolate parameters for different tasks, ensuring there is limited interference between different subsets of modules within a model \cite{rusu2016progressive, aljundi2017expert, rebuffi2017icarl}. And, replay-based approaches retain examples from past data in memory and replay these examples when the model is learning on new tasks, to alleviate the issue of catastrophic forgetting \cite{lopez2017gradient, rolnick2018experience, chaudhry2018efficient}. Other methods include applying generative models to generate pseudo data rather than replaying in memory examples \cite{shin2017continual, lavda2018continual}.


Recently, there has been a number of works that have explored applying continual learning for the task of automatic speech recognition. Sadhu and Hermansky \cite{sadhu2020continual} trained an HMM-DNN model on a sequence of four different tasks, across the Wall Street Journal (WSJ), Reverb, Librispeech, and Chime4 corpora. Chang et al. \cite{chang2021towards} built an end-to-end ASR model trained first with WSJ, then Librispeech, and finally the Switchboard corpus, while evaluating the performance of the model across the different speech recognition tasks after each update. Rather than focusing on the application of continual learning on a single task, these prior works focused on the challenge of how to update a model to perform well on new speech recognition tasks over time while retaining good performance on previously observed tasks. Both approaches require explicit information of a task boundary in the training data and as the distribution of the data from different tasks may vary dramatically, it is critical to know when the training data is switched to learn from one task to another. To overcome this limitation, Zeno et al. \cite{zeno2018task} proposed a Bayesian approach for continual learning without knowledge of task switch, and Mai et al. \cite{mai2021online} defined online continual learning in image classification by assuming that new classes or new instances of images may appear in online data streams.

\section{Online Continual Learning in ASR}
\label{sec:method}

\subsection{Problem Statement}
We define the online continual learning of ASR models as follows. First, we assume we have an initial model ($M_0$) that has been trained on a given dataset ($D_0$). This model is treated as a seed model on which a sequence of continual learning updates is applied. Second, we have a set of labeled datasets $\mathcal{D}=\{D_i\}_{i=1}^N$ which become available sequentially over time for model training. $N$ is the total number of subsets and the size of $D_i$ is in general significantly smaller than $D_0$. 
Retraining the ASR model from scratch on $\{D_i\}_{i=0}^n$ each time a new dataset $D_n$ becomes available requires significant computational cost, especially when the amount of data in $D_0$ is large and the amount of data in each $D_i$ is small.

For online continual learning of ASR models, at each time-step $n$, a well-performing method would:
\begin{itemize}
    \itemsep-0.1em
    \item Obtain better (or not degrade) accuracy compared to the previous model ($M_{n-1}$);
    \item Obtain better accuracy compared to continually training the previous model ($M_{n-1}$) on only the new data $D_n$;
    \item Obtain comparable accuracy to a model trained from scratch on all the data available at this time $\{D_i\}_{i=0}^{n}$;
    \item Require significantly less computational cost compared to training a model from scratch on $\{D_i\}_{i=0}^{n}$.
\end{itemize}

\subsection{Continual Learning with Gradient Episodic Memory}

We base our method on Gradient Episodic Memory (GEM) \cite{lopez2017gradient}. GEM is a replay-based continual learning method, which stores samples from past data as its memory. When the model encounters the data of a new task, it requires the minimization of the L2 distance between the gradients of the new data and old data, i.e.,

\vspace{-5mm}
\begin{gather}
\min_\vomega{\frac{1}{2}||\vomega - \vg||_2^2} \nonumber \\
\text{s.t. } \langle \vomega,\tilde{\vg}_i \rangle \geq 0, \forall i\in (0,\dots,n-1).
\label{eq:1}
\end{gather}


\noindent
where $\vg, \tilde{\vg}_i \in \mathbb{R}^{|\vtheta|}$ and $|\vtheta|$ is the number of parameters of the model. $n$ is the number of previous tasks and $(0,\dots,n-1)$ represents the sequence of tasks. $\vg$ is the gradient over the new data in the current task and $\tilde{\vg}_i$ is the gradient for the sampled data from $i$th task in the past, and it assumes that the positive inner product between the gradients will prevent catastrophic forgetting. $\vomega \in \mathbb{R}^{|\vtheta|}$ is the target gradient that we want to solve. This quadratic programming problem (\ref{eq:1}) is then transformed into its dual form,

\begin{equation}
\begin{aligned}
& \min_\vv{\frac{1}{2}\vv^\top \vG\vG^\top \vv + \vg^\top \vG^\top \vv} \\
& \ \ \ \ \ \text{s.t. } \vv \geq \mathbf{0}.
\end{aligned}
\end{equation}

\noindent
where $\vG=(\tilde{\vg}_1, ..., \tilde{\vg}_{n-1})$, and $\vv \in \mathbb{R}^{|\vtheta|}$. The solution of the primal form could be recovered by $\vomega = \vG^\top \vv + \vg$. After that, $\vomega$ will be used as the final gradient to update the parameters in the network.

\begin{table}[]
    \centering
    \begin{tabular}{c c c c}
        \toprule
        Characteristics & v1 & v2 & v3  \\
        \midrule
        Duration of speech & 118h & 207h & 452h \\
        No. of unique speakers & 666 & 1,242 & 2,028 \\
        No. of words & 1.7M & 2.6M & 4.9M \\
        \bottomrule
    \end{tabular}
    \caption{The overall statistics of TED-LIUM corpus v1 to v3.}
    \label{tab:dataset}
    \vspace{-5mm}
\end{table}

\vspace{-0.5mm}

\subsection{Selective Strategy}

Since we do not define clear task boundaries in our online continual learning scenario, we cannot directly follow the original training protocol directly which requires calculating gradients for all previous tasks and samples from each of the previous tasks. Instead, we perform random sampling and selective sampling respectively to sample from the memory. For the selective strategy, we assign a probability score $c_i=1-\frac{\langle \vomega,\tilde{\vg}_i \rangle}{|| \vomega|| \cdot ||\tilde{\vg}_i ||}$ to the $i$th sample, and use the weighted scores $\{c_i\}_{i=1}^{|\mathcal{M}|}$ to sample from the memory buffer $\mathcal{M}$. Such a selective strategy would select samples with less similarity to other samples in the memory more frequently, inspired by the fact that unexpected events play an important role in replayed experiences from the neuroscience perspective \cite{isele2018selective} and the gradient-based sample selection \cite{aljundi2019gradient}. Since we have a higher probability to select the samples with lower cosine similarity scores, we are relaxing the constraints in the original primal form of GEM. In other words, instead of the strict constraint that the direction of the gradients of old and new data must be similar, we release the constraint and develop the quadratic form in eq. \ref{eq:1} so that it can be rewritten as

\begin{equation}
\begin{aligned}
& \ \ \ \ \ \min_\vomega{\frac{1}{2}||\vomega - \vg||_2^2} \\
&\ \text{s.t. } \langle \vomega,\tilde{\vomega} \rangle \geq -\xi, \xi \geq 0.
\label{eq:3}
\end{aligned}
\end{equation}


\noindent
where $\tilde{\vomega}\in\mathbb{R}^{|\vtheta|}$ is the gradient of the selected memory and $\xi$ is the slack variable. The basic idea is similar to the soft gradient constraint in \cite{chen2019revisiting}. Note that since we do not assume the explicit definition of task boundaries in our case, with Karush–Kuhn–Tucker (KKT) theorem, the dual form of the quadratic programming can be written as


\begin{equation}
\begin{aligned}
& \min_v{\frac{1}{2}v^2 \tilde{\vomega}^\top \tilde{\vomega} + \vg^\top \tilde{\vomega} v} \\
&\ \ \ \ \  \text{s.t. } v \geq 0.
\end{aligned}
\end{equation}

\noindent
where $v$ is a scalar, which is different from the original dual form. Then the updated gradient can be calculated by $\vomega = v\tilde{\vomega} + \vg$ to optimize the current model.


\section{Experimental Results}
\label{sec:exp}

\subsection{Datasets}
In this work, we use TED-LIUM 1, 2 and 3 \cite{rousseau2012ted, rousseau2014enhancing, hernandez2018ted} as the datasets for the experimental evaluation. TED-LIUM is a series of datasets that consist of audios and transcripts extracted from the official TED talk website. Such property benefits us by providing audios from a variety of speakers and their captions with a wide range of topics. To be more specific, training across different topics of talks will help us demonstrate and evaluate the effectiveness of our continual learning settings, which is a more realistic setting in the real world. Meanwhile, in the series of TED-LIUM corpus, each succeeding version is a comprehensive expansion and evolution of the previous one. These are the important reasons why we choose TED-LIUM as the target dataset, since it somewhat imitates the growing data over time in real-world scenarios without explicitly defined boundaries. For simplicity, we name the difference set between TED-LIUM 1 and 2 as TED-LIUM 2-1, and similar for TED-LIUM 3-2. Detailed statistics are shown in Table \ref{tab:dataset}.

The overall comparison between the statistics of TED-LIUM 1, 2 and 3 is summarized in Table \ref{tab:dataset}. From the table, we can observe that TED-LIUM 2 and 3 expanded the original corpus by adding more acoustic and textual contents. In our experiments, we will follow the setting that the model is first pretrained on TED-LIUM 1, and then the continual learning is performed on TED-LIUM 2-1 and 3-2 with different metrics successively.

\subsection{Experiment Settings}
As the preprocessing step, we use a three-fold speed perturbation with factors 0.9, 1.0, and 1.1. The tokenization type is Byte-Pair Encoding (BPE), and we remove all the existing unknown tokens in the raw TED-LIUM corpus. All the results provided are performed with transformers \cite{karita2019comparative} and do not use language models in the decoding process for simplicity.

\begin{figure}[h]
\centering
\begin{subfigure}[b]{0.4\textwidth}
    \vspace{-3mm}
    \hspace{-5mm}
    \centering
    \includegraphics[scale=0.32]{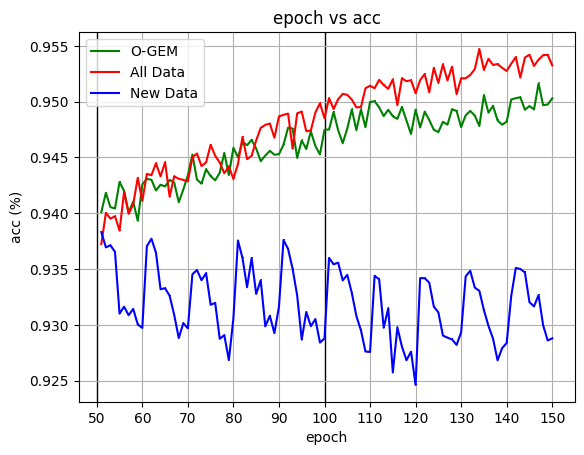}
    \vspace{-2mm}
    \caption{Validation accuracy vs. epochs}
    \label{fig:res1}
\end{subfigure}
\begin{subfigure}[b]{0.4\textwidth}
    \centering
    \includegraphics[scale=0.32]{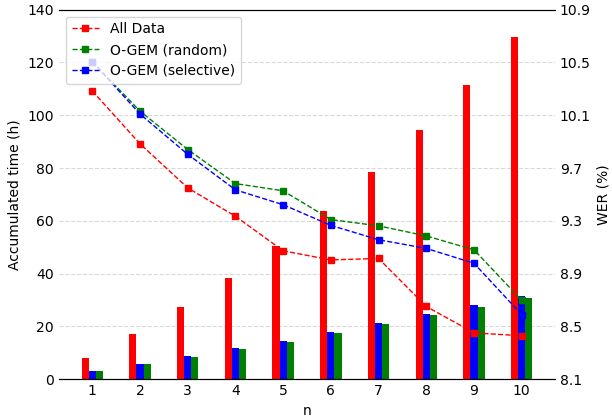}
    \caption{WER \& computation time vs. $\{D_n\}_{n=1}^N$}
    \label{fig:res2}
\end{subfigure}
    \caption{Results for validation accuracy (\%), word error rate (WER, \%) and accumulated computation time (hrs) for different methods where $N=10$.}
    \label{fig:res}
    \vspace{-5mm}
\end{figure}

First, we use TED-LIUM 1 as the baseline dataset $D_0$, as introduced in Section \ref{sec:method}, to train for 50 epochs. The final WER achieves the state-of-the-art performance of $11.0\%$ on TED-LIUM 1. After that, we randomly split TED-LIUM 2-1 and 3-2 evenly into a sequence of subsets $\{D_i\}_{i=1}^N$, and continuously train the model on the $N$ splits in order, with 10 epochs for each split. WER and the execution time will be recorded at the end of each training step, and the results are shown in Fig. \ref{fig:res}. Note that the provided validation dataset is identical for all experiments related to TED-LIUM 1, 2 and 3, so that we could compare the results of different stages fairly during the continual learning process.


\begin{table}
    \centering
    \begin{tabular}{c c c}
        \toprule
        Method & \makecell{WER (\%) after \\ TED-LIUM 2-1} & \makecell{WER (\%) after \\ TED-LIUM 3-2} \\
        \midrule
        \textit{All Data} & 9.0 & 8.1 \\
        \textit{New Data} & 10.4 & 9.8 \\
        \midrule
        EWC \cite{kirkpatrick2017overcoming} & 9.9 & 9.5 \\
        MAS \cite{aljundi2018memory} & 9.8 & 9.5 \\
        O-GEM ($N=2$) & 9.4 & 8.5 \\
        O-GEM ($N=10$) & 9.5 & 8.6 \\
        \bottomrule
    \end{tabular}
    \caption{WER results for different methods in the proposed online continual learning setting.}
    \label{tab:wer_others}
    \vspace{-10mm}
\end{table}

\subsection{Comparison to other methods}

We also compared the results of our method, the online GEM (O-GEM) on TED-LIUM corpus to other continual learning algorithms, including Elastic Weight Consolidation (EWC, \cite{kirkpatrick2017overcoming}) and Memory Aware Synapses (MAS, \cite{aljundi2018memory}). Both EWC and MAS are regularization continual learning approaches as introduced in Section \ref{sec:related}. Basically, EWC is regularizing the parameter importance based on the gradients of the loss function, while MAS controls the parameter importance from the gradients of the learned network output function. We show the comparison between the word error rate of the online continual learning with equal number of epochs for different methods in Table \ref{tab:wer_others}. The \textit{All Data} method is trained from scratch with all previous data each time when the model receives a new subset of data $D_n$, which is our upper bound performance. In contrast, O-GEM and the \textit{New Data} method are only trained on the specific new set $D_n$ each time, and O-GEM benefits from the extracted samples of the episodic memory. We will follow the names in all the discussions later. From the figure, we can conclude that EWC and MAS could decrease the WER compared to training with only new data, while O-GEM obtains a much better performance and is closer to the WER of training with all data. 


\subsection{Results Analysis}
\label{sec:analysis}

We compare the experimental results of three methods in Fig \ref{fig:res}: the O-GEM, \textit{All Data} and \textit{New Data}. All three methods are training on a sequence of subsets $\{D_i\}_{i=1}^N$, as introduced in Section \ref{sec:method}, from TED-LIUM 2-1 and TED-LIUM 3-2. 
Figure \ref{fig:res1} shows the learning curve of validation accuracy of the three methods. The training starts from epoch 50 which is the number of training epochs for the baseline model. In the \textit{New Data} scenario, we load the previous model trained on $D_{n-1}$ and continually train the model on $D_n$ with the same training strategy as the baseline model. From Figure \ref{fig:res1}, we can observe that the validation accuracy of \textit{New Data} goes up sharply at the start of each 10 epochs while goes down during the overall process of learning on the single set of data, which indicates that the model will learn new knowledge as soon as they get access to new data, but forget what it has already learned in the past quickly. Meanwhile, the performance of \textit{New Data} is stochastic across the 10 splits because its performance is largely dependent on the content of each specific new data set. The red curve is the best performing one which increases constantly, as it is retraining from scratch with both old and new data each time whenever it absorbs unseen data. Without the time-consuming retraining, however, O-GEM can achieve competitive validation accuracy in comparison to the \textit{All Data} case, while maintaining its increasing accuracy rather than causing catastrophic forgetting in the \textit{New Data} method.

Figure \ref{fig:res2} compares the WER results of selective O-GEM and \textit{All Data}. Each square marker represents the WER when the continual learning on one subset $D_n$ is finished. The \textit{All Data} training requires the longest execution time, as it requires to be retrained every time when a new $D_n$ is incoming. In contrast, our method could obtain a decreasing WER when it accesses more data over time, and have a close WER to the \textit{All Data} scenario ($8.6\%$ vs. $8.4\%$) with more than four times less computation time for both random sampling and selective sampling, and with similar computation time, selective sampling can have a further improvement of WER compared to random sampling. Note that the computation time of O-GEM and \textit{New Data} training is proportional to the amount of the new data only, while the training time of \textit{All Data} will increasingly scale up since the size of old data is growing. In other words, as the model receives more information, it will be increasingly more impractical for retraining with all data since the ratio of old data size to new data size is becoming higher over time and the computation cost of retraining is more expensive compared to training on only new data. We demonstrate the experimental results with different numbers of splits $N$ in Table \ref{tab:smallscale}, where we can observe that with more splits of new data, the ratio of computation time for O-GEM to the estimated computation time for training with all data is becoming smaller. The reason is that \textit{All Data} would have to retrain the model from scratch for every $D_n$, while our method could save more computation time, which is closer to the real-world problem setting especially when $N$ is large.

\begin{table}[h]
    \vspace{-3mm}
    \centering
    \begin{tabular}{c c c c}
        \toprule
        $N$ & \makecell{Avg. hours \\ per split} & \makecell{Model update time \\ diff. vs. \textit{All Data}} & Final WER (\%)  \\
        \midrule
        0 & - & - & 11.0 \\
        2 & 167.0 & 79.05\% &  8.5 \\
        10 & 33.4 & 23.80\% &  8.6 \\
        20 & 17.3 & 13.04\% & 8.9 \\
        100 & 3.3 & 2.86\% & 9.2 \\
        200 & 1.7 & 1.92\% & 9.5 \\
        \bottomrule
    \end{tabular}
    \caption{Results for the O-GEM experiments with updates of different numbers of splits $N$. The time for \textit{All Data} is estimated when $N>10$ since it is impractical. Note that $N=0$ refers to the starting point of the online continual learning, and $N=2$ means no splitting is performed on TED-LIUM 2-1 \& 3-2.}
    \label{tab:smallscale}
    \vspace{-7mm}
\end{table}

\subsection{Compatibility with Self-Supervised Learning}

\begin{figure}[h]
    \centering
    \includegraphics[scale=0.32]{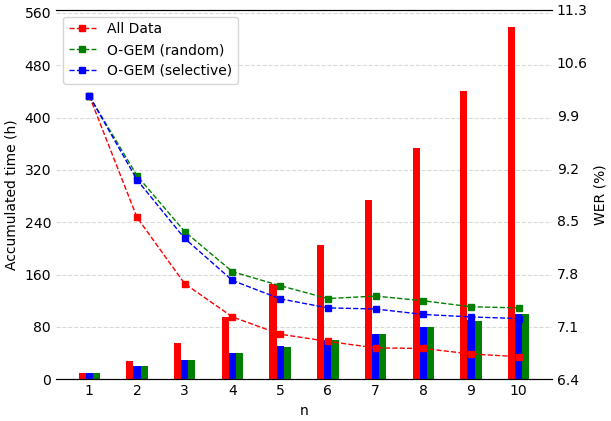}
    \vspace{-3mm}
    \caption{WER \& computation time vs. $\{D_n\}_{n=1}^N$ with HuBERT as the self-supervised learning frontend.}
    \label{fig:hubert}
\end{figure}

Self-supervised learning has achieved great success in a variety of fields, benefiting many downstream tasks from learning effective representations from data. Especially in automatic speech recognition, many methods have been proposed to pretrain models by using a large amount of untranscribed data to learn speech features effectively. Therefore, we also use Figure \ref{fig:hubert} to demonstrate the compatibility of our method with one of the latest self-supervised learning models, HuBERT \cite{hsu2021hubert}, which pretrains the representation model by classification tasks using pseudo-labels from the data. We can observe that with HuBERT as the frontend, the WER curve converges faster in all three scenarios, while keeping our previous analysis consistent, which indicates the compatibility of our models with self-supervised learning frontends.

\vspace{-2mm}
\subsection{Compatibility with other datasets}

In this section, we also explore the compatibility of our continual learning methodology on other datasets, the Wall Street Journal (WSJ) corpus. We choose this dataset because it also consists of two successive versions, WSJ0 and WSJ1, which would be similar to be adapted into our proposed continual learning setting. From Table \ref{tab:wsj}, we can see that our method could achieve a close WER to retraining all data from scratch, while saving around four times less computation time, which is consistent with our prior analysis on the TED-LIUM corpus, and indicates our compatibility with other datasets as well.
\begin{table}
    \centering
    \begin{tabular}{c c c}
    \toprule
       N & \makecell{Model update time \\ diff. vs. \textit{All Data}}  & WER (\%) \\
    \midrule
       \textit{All Data} & - & 4.9 \\
       \textit{New Data} & 72.35\% & 6.7 \\
    \midrule
       O-GEM ($N=2$)  & 78.76\% & 5.4 \\
       O-GEM ($N=10$)  & 25.37\% & 5.6 \\
    \bottomrule
    \end{tabular}
    \caption{Experimental results for WSJ0-1.}
    \label{tab:wsj}
    \vspace{-9mm}
\end{table}

\vspace{-2mm}
\section{Conclusion}
\label{sec:conc}

In this paper, we proposed a novel experimental setting for online continual learning in automatic speech recognition, focusing on the case where additional training data for the same speech recognition task becomes available incrementally over time. We demonstrate that by performing incremental model updates with an online Gradient Episodic Memory (GEM) method and a selective sampling strategy, we can achieve competitive performance to retraining an ASR model from scratch each time the model has access to new data, while requiring significantly lower computation cost. We also demonstrate that our online GEM outperforms other continual learning methods under the same setting, as well as its compatibility with self-supervised learning models and other datasets. We believe that this work provides a first exploration into online continual learning for end-to-end speech recognition models, and makes a step towards realizing Lifelong Learning for speech interfaces.



\bibliographystyle{IEEEtran}

\bibliography{mybib}


\end{document}